\documentclass[letterpaper, 10pt, conference]{ieeeconf} 

\IEEEoverridecommandlockouts 
\usepackage[utf8]{inputenc}

\usepackage{amsmath} 
\usepackage{textcomp}
\usepackage{gensymb}
\usepackage{comment}
\usepackage{todonotes}
\usepackage{graphicx}
\usepackage{floatrow}
\usepackage{url}
\usepackage{dblfloatfix}    

\title{A shallow residual neural network to predict the visual cortex response}

\author{\authorblockN{Anne-Ruth Jos\'e Meijer}\authorblockA{Universiteit van Amsterdam, The Netherlands} \and \authorblockN{Arnoud Visser}\authorblockA{Universiteit van Amsterdam, The Netherlands} 
}

\begin{document}

\maketitle
\thispagestyle{empty}
\pagestyle{empty}

\begin{abstract}
Understanding how the visual cortex of the human brain really works is still an open problem for science today. A better understanding of natural intelligence could also benefit  object-recognition algorithms based on convolutional neural networks. In this paper we demonstrate the asset of using a shallow residual neural network for this task. The benefit of this approach is that earlier stages of the network can be accurately trained, which allows us to add more layers at the earlier stage. With this additional layer the prediction of the visual brain activity improves from $10.4\%$ (block 1) to $15.53\%$ (last fully connected layer). By training the network for more than 10 epochs this improvement can become even larger.
\end{abstract}

\section{Introduction}

Current state-of-the-art convolutional neural networks (CNNs) incorporate operations like maximum-based pooling of inputs  \cite{yamins2016using}, which were directly inspired by single-cell recordings from the V1 region of the mammalian visual cortex \cite{hubel1959receptive}. 
This inspiration is also beneficial for object recognition, because there is a correlation  \cite{yamins2014performance} observed between a CNNs ImageNet's  performance \cite{SCHMIDHUBER201585} and its Brain-Score \cite{brainscore}. However, this 
correlation can no longer be found for the
most advanced CNNs \cite{yamins2014performance}. Based on the hypothesis that by simplifying CNNs one could improve the understanding of ventral stream in the visual cortex, this year's challenge of the Algonauts project is introduced \cite{algonauts}. The target of the 2019 challenge is to predict the activity in the human visual brain responsible for object recognition in two regions; the early visual cortex (EVC) and the inferior temporal (IT) cortex. Our study seems to confirm that simpler CNNs (more shallow) can have competitive prediction power when compared to "deeper" CNNs, because the earlier layers can be trained faster and more accurately.

\section{Explaining the Human Visual Brain Challenge}

\begin{figure}[ht]
    \centering
    \includegraphics[height=0.11\textheight]{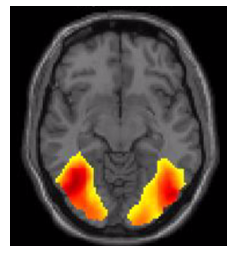} %
    \includegraphics[height=0.12\textheight]{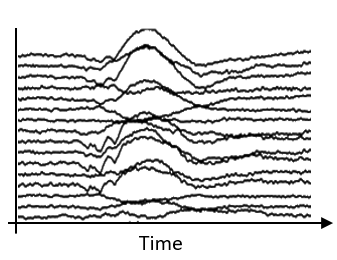} %
    \caption{Brain activity recorded with respectively the fMRI and MEG technique (Courtesy Algonauts project\protect\footnotemark)}
    \label{fig:recordings}
\end{figure}

The goal of the 2019 challenge is predict the response of two parts of the human brain responsible for object recognition. Two datasets are provided of brain recordings of 15 human subjects looking at pictures with objects from the ImageNet dataset \cite{deng2009imagenet}. The brain recordings are respectively functional magnetic resonance imaging (fMRI) for Track 1 and Magnetoencephalography (MEG) for Track 2 (see Fig.~\ref{fig:recordings}). fMRI is a technique which detect changes in blood flow with spatial resolution of millimeters; MEG is a technique that changes in magnetic fields with a temporal resolution of milliseconds.

\footnotetext{\url{http://algonauts.csail.mit.edu/fmri_and_meg.html}}

\begin{figure}[hb]
    \centering
    \includegraphics[height=0.31\textheight]{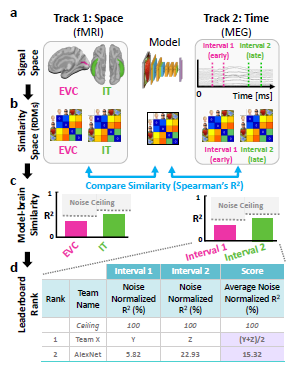} 
        \caption{The evaluation procedure of the 2019 Challenge (Courtesy \cite{algonauts})}
    \label{fig:procedure}
\end{figure}

The datasets are recorded in such a way that the observations correspond with activity in the early visual cortex (EVC) an the inferior temporal (IT) cortex, with respect to space (fMRI-data of Track 1) and time (MEG-data of Track 2). The response of the human brain and the CNN models is made possible by converting the incommensurate signals into the same similarity space, which are defined by representational dissimilarity matrices (RDMs) \cite{kriegeskorte2013representational}.
The recorded and predicted RDM are than compared based on the Spearman correlation, which is defined as the Pearson correlation between rank variables \cite{spear}. The result is normalized against the correlation an ideal model could give, taking into account the variation and noise in the dataset. As baseline, the prediction of the classic CNN AlexNet is given \cite{alex}. The overall evaluation procedure is illustrated in Fig.~\ref{fig:procedure}.

\begin{figure*}[t] 
    \centering
    \includegraphics[width=1\textwidth]{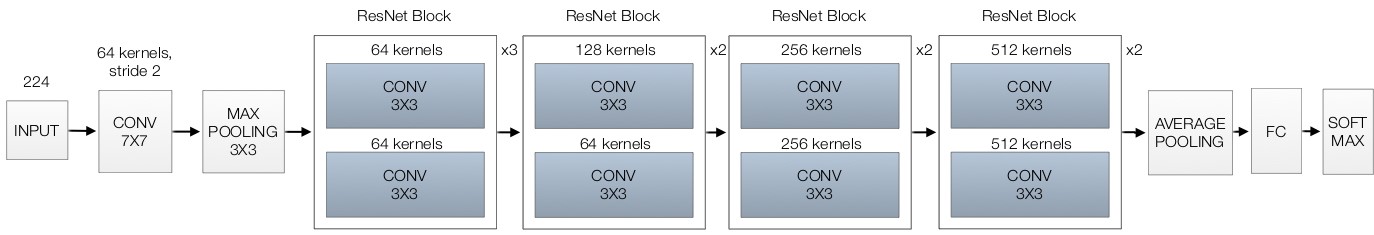}\\
    \caption{ResNet20 architecture}
    \label{fig:resnet19}
\end{figure*}

\section{Related Work}

The challenge is inspired by the initiative to find a Brain-Score \cite{brainscore}, which found a correlation between the ImageNet performance and the Brain-Score. Yet, for the CNNs with the highest performance this correlation becomes less strong. The conclusion of the study was that DenseNet169, CORnet-S and ResNet-101 were the most brain-liked CNNs. Yet, a number of  smaller (i.e more shallow) networks performed quite competitive, leaving the road open to better understand the ventral stream with simpler CNNs. 

Interesting here is the good performance of CORnet-S \cite{cornets}, which is inspired by 
the relatively shallow cortical hierarchy (4-8 levels) which is observed in human brains. CORnet-S contains elements of ResNet and DenseNet, in the sense that it allows the backpropagation to reach the earlier layers by skipping layers, yet it does this with a biological type of unrolling. Unfortunately we could not reproduce the good performance of CORnet-S for this challenge (see for more details \cite{Meijer2019thesis}), so this biologically interesting approach is no further studied in this paper.

Another interesting approach are the deep CNNs proposed by Kar \textit{et al} \cite{kar2019evidence}. Here they observed that deeper CNNs predicted neural responses better than shallower models, if they have unrolling mechanisms present. The best predictions were from ResNet-50 and ResNet-101. Yet the prediction were for the MEG-data of the IT region, while in this study we want to concentrate on the fMRI-data of both regions.

\section{Replication study}

To get a feeling on the challenge, a number of networks which scored well at the Brain-Score leaderboard\footnote{\url{http://www.brain-score.org/}} were tested on the challenge dataset. This data and metrics are not completely the same, so the results can not 1-to-1 be translated. The tested convolutional neural networks are AlexNet, VGG,  ResNet18, ResNet-34, ResNet-50, ResNet-101, ResNet-152, DenseNet121, DenseNet161, DenseNet169, DenseNet201, CorNet-S, and CorNet-Z. AlexNet is used as baseline. All networks were already pre-trained on the ImageNet dataset.

\begin{figure}[H]
    \centering
    \includegraphics[width=1\textwidth]{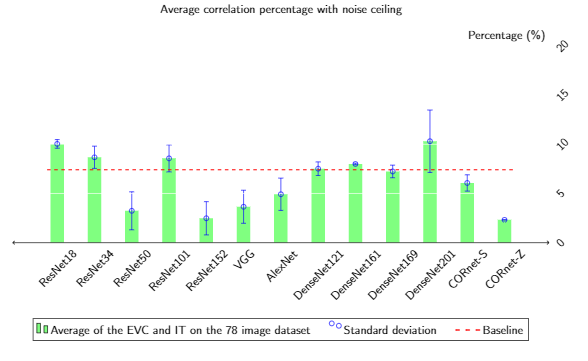}\\
    \caption{The noise normalized squared Spearman correlation percentage of the Algonauts test set for different CNN's, along with the standard deviation.}
    \label{fig:barplot2}
\end{figure}

Each network was tested on the provided 92 and 118 datasets. The best layer of a network is the layer that scored average the best on both training datasets. The best layers of the pre-trained networks were submitted to the Algonauts challenge. In figure \ref{fig:barplot2} the score of the submitted models at the Algonauts challenge is visualized, more details can be found in \cite{Meijer2019thesis}. The score is the average noise normalized squared correlation score of the EVC and IT region for each network. The average highest scoring network is the DenseNet201, with second place ResNet-18 and third ResNet-34. After this top 3, only ResNet-101, DenseNet121 and DenseNet161 score higher than the challenge baseline.

Based on the good performance of the 'shallow' ResNet-18 and ResNet-34, we decided to design an own 'shallow' network (ResNet-20) and train this for the competition.

\section{Our method}

For this study a variation of a ResNet with 20 layers is designed. Because our hypothesis is that the earliest layers are important, this is in principal the same architecture as ResNet-18, but the first ResNet block occurring 3 times (instead of 2). This block has the same number of layers as the ResNet blocks in the design of ResNet-18 and ResNet-34. The architecture of the ResNet-20 is visualized in figure \ref{fig:resnet19}. As can be seen, the main building blocks are so-called ResBlocks. These ResBlocks contain two convolution layers, each with a kernel of 3x3. Each of these ResBlocks can be skipped to bring the feedback signal fast back to the earlier layers.. This is only possible because the spatial size of the input of every ResBlock does not differ from the output of the block. The ResNets includes a max pooling, an average pooling, a fully connected, and a softmax layer.  

This ResNet-20 is trained on a GPU containing the AMD Radeon RX VEGA 64 Chip. The network is trained for a maximum of 120 epochs, although the first and every 5 epochs a model is saved. This ensures that interim models can be tested. Similar to the testing phase, the images used while training are normalized and re-sized. The optimizer starts with a learning rate of 0.1, a momentum of 0.9, and a weight decay of 0.0001. The learning rate affects the speed at which a network trains \cite{cnn}. A learning rate of 0.1 indicates that each time the weights are updated, they are updates with 10 percent of the estimated weight error. Momentum is a method used for pushing the gradient in the right direction \cite{cnn}. During training an optimization function can get stuck in a local minimum, while the algorithm thinks the global minimum is reached. The momentum tries to avoid this problem by increasing the steps taken towards the minimum. Within a local minimum the increase of steps could cause the function to jump out of a local minimum. The value of the momentum indicates with what size the steps increase. The weight decay is the value with which the weights are multiplied after each update \cite{cnn}, preventing the weights from growing too large. 

\section{Results}

At the time of writing the maximum of 120 epochs is not reached, models were saved after training for 1 epoch, 5 epochs and 10 epochs. All the models produce results above the baseline. In all cases the fully connected layer (FC), the last layer before the SOFTMAX, gave the best prediction.

\begin{figure}[H]
    \centering
    \includegraphics[width=1\textwidth]{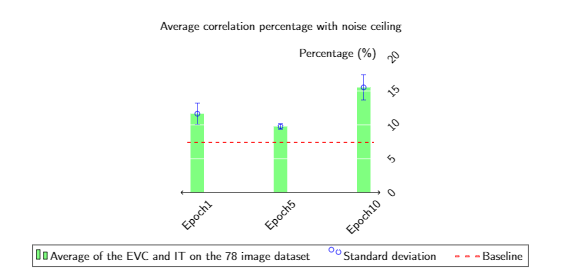}\\
    \caption{The average noise normalized squared Spearman correlation percentage of the EVC and IT of the test set for different trained ResNet-20 models, along with the standard deviation}
    \label{fig:barplot4}
\end{figure}

The response of the FC layers were all submitted to the Algonauts challenge. Figure \ref{fig:barplot4} shows these results, along with the standard deviation. The model trained with 10 epochs preforms the best, which looks promising for further training.

The result is not (yet) good enough to outperform other participants, but good enough for a top-10 ranking at the Algonauts leaderboard\footnote{\url{http://algonauts.csail.mit.edu/challenge.html}}.

\section{Discussion \& Future Work}

This is a promising result, with a 'shallow' network as ResNet-20 outperforming deeper, more complex networks as DenseNet201 and ResNet-101. The Algonaut score of ResNet-20 is 15.53\%, which should be compared with 10.31\% of the best performing architecture (DenseNet201) in the replication scenario. Without further experiments it is difficult to conclude if this should be contributed to the 'shallow' design with extra convolutional layer in the earliest ResBlock, or to the training. Yet, DenseNet201 is already outperformed after the first epoch of training, so we should not overestimate the training contribution.

\section{Conclusion}

In this paper we demonstrate the asset of using a shallow residual neural network with 20 layers. The benefit of this approach is that earlier stages of the network can be accurately trained, which allows us to add more layers at the earlier stage. With this additional layer the prediction of the visual brain activity improves to $15.53\%$ (last fully connected layer) outperforming deep neural networks with more than 200 layers.

\addtolength{\textheight}{-12cm}   


\bibliographystyle{IEEEtran}
\bibliography{IEEEabrv,references} 

\end{document}